\documentclass[conference]{IEEEtran}
\usepackage{times}

\usepackage[numbers]{natbib}
\usepackage{multicol}
\usepackage[bookmarks=true]{hyperref}

\usepackage{graphicx}
\usepackage{multirow}
\usepackage{booktabs}

\usepackage{amsfonts}       % blackboard math symbols
\usepackage{amsmath}        % math environments
\usepackage{algorithm}      % algorithm figure environment
\usepackage{algpseudocode}  % pseudocode environment
\usepackage{listings}       % code listing package
\usepackage{xcolor}         % color highlighting
\usepackage{float}          % more control over floats
\usepackage{subcaption}     % subfigures and subcaptions

\begin{document}

\title{Solving the Baby Intuitions Benchmark \\ with a Hierarchically Bayesian Theory of Mind}

% You will get a Paper-ID when submitting a pdf file to the conference system
% \author{Author Names Omitted for Anonymous Review. Paper-ID 13}

\author{
\authorblockN{
Tan Zhi-Xuan\textsuperscript{1},
Nishad Gothoskar\textsuperscript{1},
Falk Pollok\textsuperscript{2}, \\
Dan Gutfreund\textsuperscript{2},
Joshua B. Tenenbaum\textsuperscript{1},
Vikash K. Mansinghka\textsuperscript{1}}
\authorblockA{
\textsuperscript{1}MIT \qquad
\textsuperscript{2}MIT-IBM Watson AI Lab\\
Email: xuan@mit.edu, nishadg@mit.edu, falk.pollok@ibm.edu \\
dgutfre@us.ibm.edu, jbt@mit.edu, vkm@mit.edu
}
}

% avoiding spaces at the end of the author lines is not a problem with
% conference papers because we don't use \thanks or \IEEEmembership

%Homer Simpson\authorrefmark{2},
%Montgomery Scott\authorrefmark{3} and
%Eldon Tyrell\authorrefmark{4}}
%Telephone: (800) 555--1212, Fax: (888) 555--1212}
%\authorblockA{\authorrefmark{4}Tyrell Inc., 123 Replicant Street, Los Angeles, California 90210--4321}}

\maketitle

\begin{abstract}
To facilitate the development of new models to bridge the gap between machine and human social intelligence, the recently proposed Baby Intuitions Benchmark provides a suite of tasks designed to evaluate commonsense reasoning about agents' goals and actions that even young infants exhibit. Here we present a principled Bayesian solution to this benchmark, based on a hierarchically Bayesian Theory of Mind (HBToM). By including hierarchical priors on agent goals and dispositions, inference over our HBToM model enables few-shot learning of the efficiency and preferences of an agent, which can then be used in commonsense plausibility judgements about subsequent agent behavior. This approach achieves near-perfect accuracy on most benchmark tasks, outperforming deep learning and imitation learning baselines while producing interpretable human-like inferences, demonstrating the advantages of structured Bayesian models of human social cognition.
\end{abstract}

\IEEEpeerreviewmaketitle

\section{Introduction}
\label{sec:intro}

From a very young age, humans display remarkable psychological intuition about the mental lives and likely behavior of other agents. For example, developmental psychologists have shown that young infants expect agents to have object-centered goals \cite{gergely1995taking,woodward1999infants}, to act efficiently to achieve those goals \cite{gergely1997teleological,liu2017six}, to choose goals based on inferred preferences \cite{repacholi1997early,buresh2007infants}, and to undertake instrumental actions in pursuit of those goals \cite{hernik2015infants,gerson2015shifting}. In contrast, many contemporary machine learning approaches do not appear to demonstrate this structured prior knowledge, leading to inferences and predictions about agents that correspond poorly with human intuition \cite{shu2021agent,gandhi2021baby}.

To facilitate the bridging of this gap, \citet{gandhi2021baby} recently published the Baby Intuitions Benchmark (BIB), a suite of tasks designed to test the ability of computational models to judge the plausibility of agent behavior in accordance with infant psychological intuitions. In each task instance, the model is presented with a series of familiarization trials that contain information about an agent's goals and dispositions, followed by a test trial that may be consistent or inconsistent with those goals and dispositions. They show that deep learning and reinforcement learning baselines fail to perform well on these tasks, leaving it an open question as to what kinds of modeling approaches might succeed.

In this paper we describe a principled Bayesian approach to solving this benchmark, which we refer to as a hierarchically Bayesian Theory of Mind (HBToM). Building upon research on action understanding as Bayesian inverse planning \cite{shu2021agent,baker2009action,jara2019naive,zhi2020online,alanqary2021modeling}, we extend previous Bayesian Theory of Mind models with hierarchical priors over the efficiency and preferences of agents. This allows the HBToM model to accumulate evidence about an agent's efficiency and preferred goals across multiple observation episodes, thereby forming expectations about future agent goals and behavior from a small number of previous interactions. The model can then be used to quantify the plausibility of subsequent observations based on how future inferences and behavior deviates from those expectations. We show that this approach achieves near-perfect accuracy on most benchmark tasks, while also producing interpretable features for its plausibility judgements which correspond to changes in attributed mental states and dispositions, demonstrating the practical and theoretical advantages of structured Bayesian models of human social cognition.

\section{The Baby Intuitions Benchmark}
\label{sec:bib}

We briefly review the format of the Baby Intuitions Benchmark to aid understanding of the rest of this paper. BIB consists of a suite of five task sets, evaluating the ability of a computational model to account for an aspect of infant intuitive psychology:

\begin{enumerate}
    \item \textbf{Efficiency}: Agents are understood to act efficiently to reach their goals, unless evidence suggests otherwise.
    \item \textbf{Preference}: Agents are understood to prefer object-based goals that they have sought in the past.
    \item \textbf{Multi-Agent}: Separate agents are understood to have separate object preferences.
    \item \textbf{Inaccessible Goals}: Agents are understood to seek out an accessible goal over an inaccessible one.
    \item \textbf{Instrumental Actions}: Agents are understood to take instrumental actions to achieve their goals.
\end{enumerate}

Each task set consists a large number of episodes, where each episode consists of 8 familiarization trials demonstrating agent behavior, followed by a test trial that may contain plausible or implausible behavior given what was shown in the familiarization trials. For example, in the preference task, an episode might contain 8 trials of an agent seeking a blue object instead of a red object, but in the test trial, the agent seeks out the red object instead. To succeed on the benchmark, models must accurately judge whether the test trial is plausible or implausible, measured relative to a paired episode with the same set of familiarization trials but a different test trial (or vice versa). Full benchmark details can be found in \cite{gandhi2021baby}.

\section{A Hierarchically Bayesian Theory of Mind}
\label{sec:hbtom}

\begin{figure}
    \centering
    \includegraphics[width=\columnwidth]{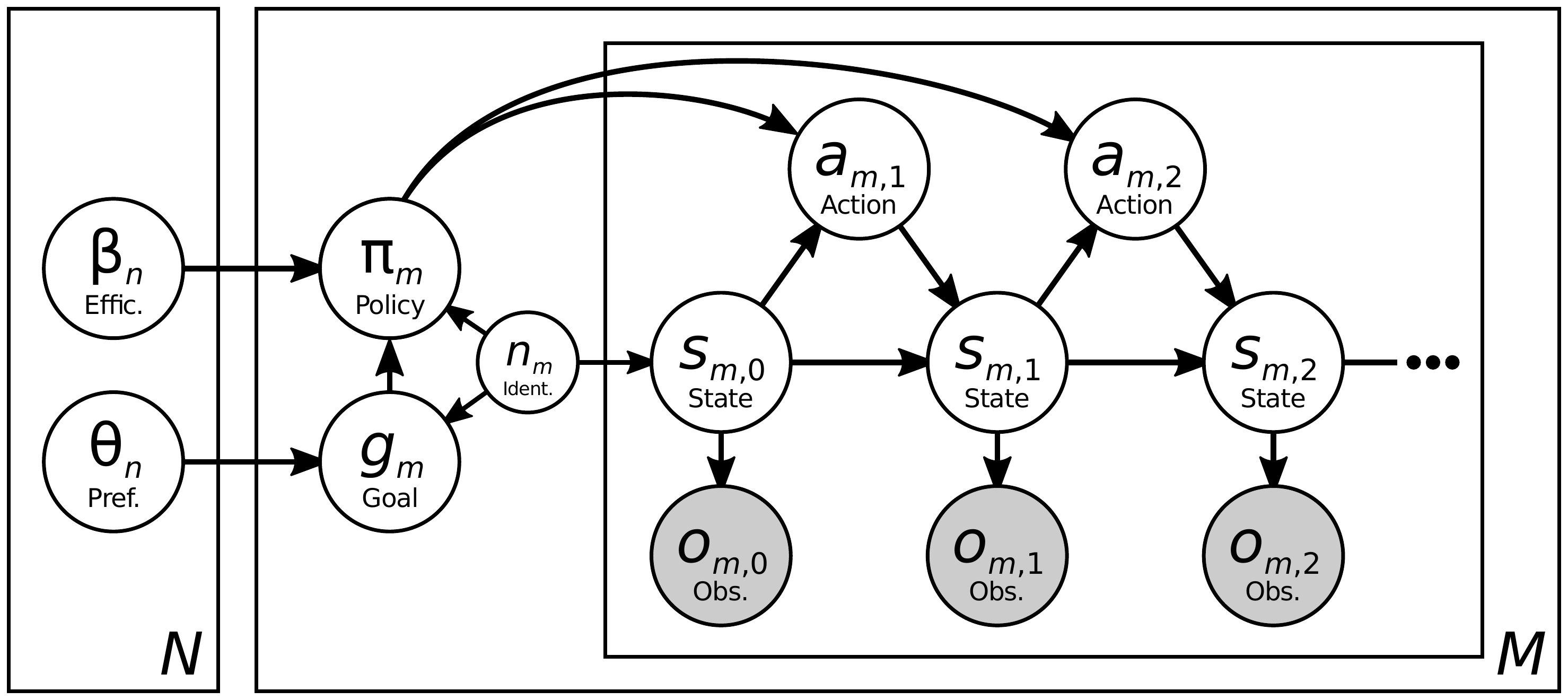}
    \caption{Graphical model of our hierarchical Bayesian Theory of Mind, where $N$ is the number of agents and $M$ the number of the trials.}
    \label{fig:graphical-model}
\end{figure}

We now introduce our hierarchically Bayesian Theory of Mind, a hierarchical Bayesian model of $N$ agents acting across $M$ viewing trials. The structure of the model is represented by Equations \ref{eq:preference-prior}--\ref{eq:observation-noise}, and is depicted graphically in Figure \ref{fig:graphical-model}. For each agent $n$, we model uncertainty over which object it tends to prefer, represented as a probability vector $\theta_n$. We also model uncertainty over the efficiency $\beta_n$ of the agent, with higher $\beta_n$ corresponding to more direct motion towards the goal. For each trial $m$, there is an associated agent $n_m$ and corresponding initial state $s_{m,0}$. The agent's goal $g_m$ for that trial is distributed according to their preference probabilities $\theta_{n_m}$. Given this goal $g_m$, the agent forms a policy $\pi_m$ to achieve the goal, modulated by its efficiency $\beta_{n_m}$. Finally, the agent acts according to this policy, leading to a state-action trajectory $(s_{m,0}, a_{m,1}, s_{m,1}, ..., a_{m,T}, s_{m,T})$ and the observer receives observations $(o_{m,0}, ..., o_{m,T})$ of the states. 

\begin{alignat}{2}
\textbf{For each agent $n$:} & \nonumber \\
\textit{Preference prior:}& \qquad \theta_n \sim P(\theta_n) \label{eq:preference-prior} \\
\textit{Efficiency prior:}& \qquad \beta_n \sim P(\beta_n) \label{eq:efficiency-prior} \\
\textbf{For each trial $m$:} & \nonumber \\
\textit{Agent identity:}& \qquad n_m \sim P(n_m) \label{eq:agent-identity} \\
\textit{Goal selection:}& \qquad g_m \sim P(g_m|\theta_{n_m}) \label{eq:goal-selection} \\
\textit{Policy construction:}& \qquad \pi_m \sim P(\pi_m|g_m,\beta_{n_m}) \label{eq:policy-construction} \\
\textit{Initial state:} & \qquad s_{m,0} \sim P(s_{m,0}|n_m) \label{eq:initial-state} \\
\textbf{For each timestep $t$:} & \nonumber \\
\textit{Action selection:}& \qquad a_{m,t} \sim P(a_{m,t} | s_{m,t-1}, \pi_m) \label{eq:action-selection} \\
\textit{State transition:} & \qquad s_{m,t} \sim P(s_{m,t} | s_{m,t-1}, a_{m,t}) \label{eq:state-transition} \\
\textit{Observation noise:}& \qquad o_{m,t} \sim P(o_{m,t}|s_{m,t}) \label{eq:observation-noise}
\end{alignat}

In the following sections, we describe the details of each model component. We then explain how Bayesian inference about latent agent parameters is performed, the results of which can be used to form plausibility judgements.

\subsection{Agent Priors}

Two key psychological findings tested by BIB are that (i) human infants assume that agents have preferences, and (ii) they also assume that agents act efficiently to achieve their goals. These assumptions, however, are not wholly fixed, as evidenced by how infants habituate to the preferences and efficiency of observed agents as they act across multiple trials \cite{buresh2007infants,gergely1995taking,liu2017six}. To account for such habituation, we adopt a \emph{hierarchical} model similar to hierarchical Bayesian inverse reinforcement learning \cite{choi2014hierarchical}, placing priors on the goal probabilities $\theta_n$ and efficiency $\beta_n$ of the agent:
\begin{align*}
    P(\theta_n) &= \text{Dir}(\alpha_1, ..., \alpha_G) \\
    P(\beta_n) &= \text{Inv-Gamma}(a, b)
\end{align*}
The probabilities $\theta_n$ represent a categorical distribution over goal objects, but can also be interpreted as a soft preference: The more an agent prefers an object, the more likely it will select the object as a goal in a new trial. By watching which object an agent selects across multiple trials, an observer can infer likely values of $\theta_n$, and hence perform few-shot learning of the corresponding object preference. Similarly, by watching how optimal the agent's behavior is, the observer can infer likely values of the efficiency parameter $\beta_n$, habituating to either efficient or inefficient behavior after multiple trials.

\subsection{Goal Selection}

For scenes with multiple objects, the agent $n_m$ in each trial $m$ has to decide which object shall be its goal. By default, it does so according to its soft preference $\theta_{n_m}$:
\begin{equation*}
P(g_m|\theta_{n_m}) =  \text{Categorical}(\theta_{n_m})
\end{equation*}
However, on some trials, objects may turn out to be inaccessible. In these cases, we assume the agent is aware that some objects are out of reach, and adjusts its preferences $\theta_{n_m}$ by eliminating the inaccessible objects as possibilities, and renormalizing the remaining probabilities, giving an adjusted preference $\tilde \theta_{n_m}$. We call this operation \textsc{Feasibilize}, giving the resulting goal distribution:
\begin{align*}
\tilde \theta_{n_m} &= \textsc{Feasibilize}(\theta_{n_m}) \\
P(g_m|\theta_{n_m}) &= \text{Categorical}(\tilde \theta_{n_m})
\end{align*}
By assuming that agents ``feasibilize'' their preferences, our model accounts for the intuitive prediction that agents will reach for an accessible but dispreferred object over an inaccessible but preferred one.

\subsection{Policy Computation}

Once a goal $g_m$ has been selected, the agent plans to achieve that goal by constructing a policy $\pi_m$, specifying a distribution over goal-directed actions $a$ for each state $s$ the agent might find itself in. This can be done by representing the task of reaching the goal $g_m$ as a Markov Decision Process (MDP) with negative rewards on actions and reaching the chosen goal as a termination condition. Solving the MDP produces Q-values $Q(s, a)$ corresponding to the expected future reward of taking action $a$ in state $s$. In a deterministic environment, $Q(s,a)$ thus corresponds to the (negative) total action cost require to reach the goal $g_m$ from $s$ by taking action $a$. For an optimally efficient agent, actions are then chosen by always maximizing the Q-value. To account for potential inefficiency, however, we instead make the standard assumption that actions are sampled from a Boltzmann distribution, parameterized by the agent's efficiency $\beta_n$:

\begin{equation*}
    \pi_m(a|s) = \frac{\exp(\beta_n Q(s, a))}{\sum_{a'} \exp(\beta_n Q(s, a'))}
\end{equation*}

Higher values of $\beta_n$ lead to more goal-directed actions, with $\beta_n=\infty$ being optimal, and $\beta_n = 0$ leading to uniformly random actions. Because we model uncertainty over $\beta_n$, our model is able to account for a range of agent behaviors, ranging from highly efficient to random, and also habituate to multiple exposures to one sort of behavior or the other.

\subsection{Environment States and Observations}

We model the environment dynamics $P(s_t|a_t, s_{t-1})$ of the BIB environment as a discretized and deterministic gridworld, where at each time step $t$, the agent may move between adjacent unobstructed grid cells (including diagonal moves), pick up objects in adjacent cells, or use a key it is holding with an adjacent ``lock''. Each action is assumed to incur a unit cost, except for diagonal moves, which cost $\sqrt{2}$. These dynamics are specified using the Planning Domain Definition Language (PDDL), a commonly used specification language for symbolic planning domains and MDPs \cite{mcdermott1998pddl}.

Given an environment state $s_t$, we can specify an observation model $P(o_t|s_t)$ that describes how observations are distributed. Since the BIB dataset provides noise-free observations, the results presented here assume direct access to state variables, though a variant of our model that converts 3D observations to symbolic states exhibited similar performance on the DARPA Machine Common Sense version of the BIB dataset. Extensions to noisy image observations are also possible through Bayesian inverse graphics \cite{gothoskar20213dp3}.  

\subsection{Bayesian Inference}

Having developed our HBToM model, Bayesian inference can be performed over this model to determine an agent's likely dispositions and mental states, which can then be used to quantify the observer's expectations. Let $\vec o_{m,t} := (o_{1,1:T_1}, o_{2,1:T_2}, ... o_{m,1:t})$ be the observations received so far up to timestep $t$ of trial $m$, where $o_{1,1:T_k}$ are all observations for the $k$th trial. We are interested in inferring the efficiency $\beta_{n_m}$ of agent $n_m$, as well as their goal $g_m$ for trial $m$:
\begin{align*}
\textit{Efficiency posterior:} \qquad &P(\beta_{n_m}|\vec o_{m,t}) \\
\textit{Goal posterior:} \qquad &P(g_m|\vec o_{m,t})
\end{align*}

Computing these posteriors requires applying Bayes rule and marginalizing over all unobserved variables. While this may appear to be computationally intensive, conjugacy assumptions along with a few simplifications to the model allow us to do this tractably (details in the Appendix).

\section{Quantifying Plausibility}
\label{sec:plausibility}

\begin{figure}[h]
    \centering
    \includegraphics[width=\columnwidth]{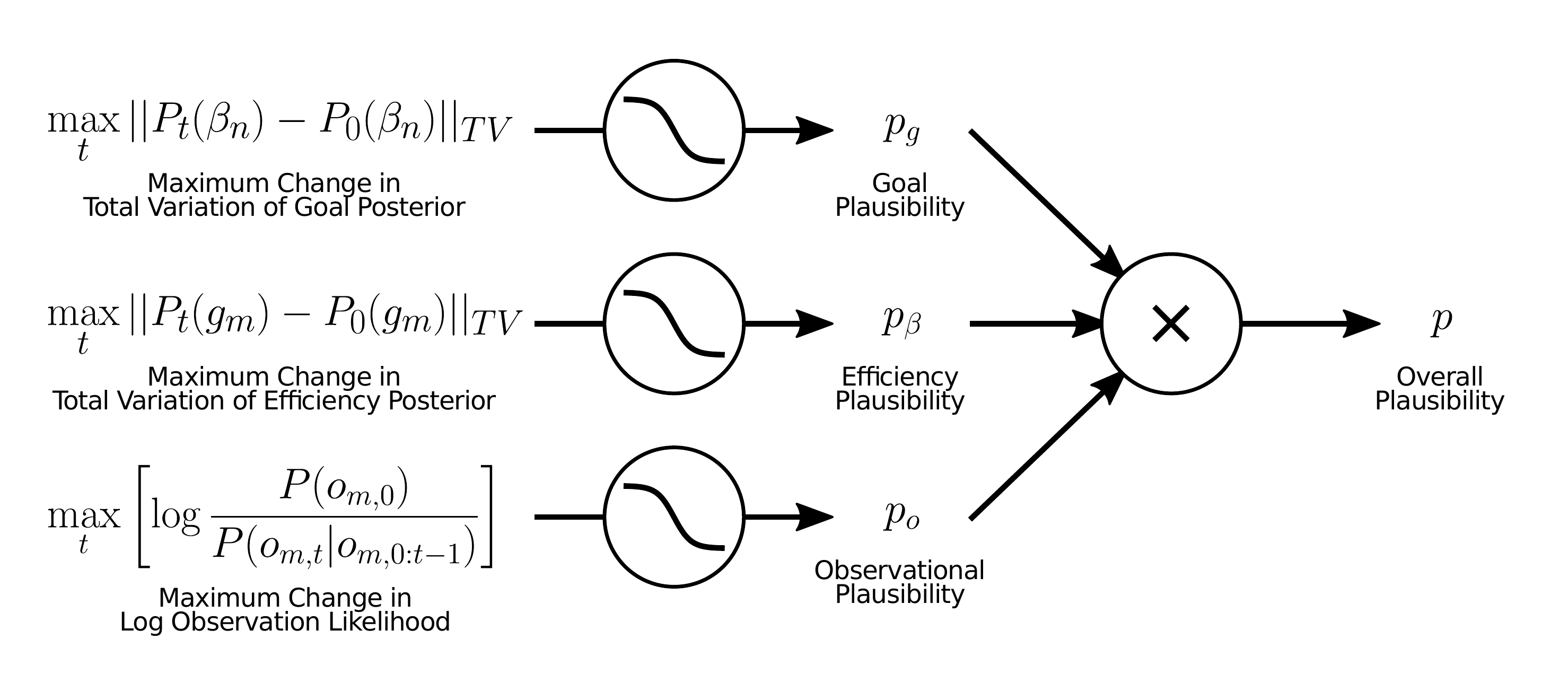}
    \caption{Classifier that quantifies the plausibility $p$ of a trial $m$ based on measures of surprise with respect to the goal $g_m$, efficiency $\beta_{n}$, and observations $o_{m,1:T}$.}
    \label{fig:plausibility-classifier}
\end{figure}

In order to compute the plausibility of a test trial, we assume that observers are simultaneously predicting and inferring the mental states and dispositions of observed agents, experiencing surprise when their inferences do not match their initial predictions. In addition, observers may be surprised when an observation just cannot be explained well by their model of the agent, given everything they have seen before. We quantify these notions of surprise using the following surprise metrics:
\begin{itemize}
    \item \textbf{Total variation in the goal posterior}:
    \[\max_t ||P_t(g_m) - P_0(g_m)||_{TV}\]
    where $P_0(g_m)$ and $P_t(g_m)$ are the inferred distributions over goals $g_m$ at the start and time $t$ of trial $m$ respectively. This captures surprise due to the actual goal differing from the one initially predicted. \\
    
    \item \textbf{Total variation in the efficiency posterior}
    \[\max_t ||P_t(\beta_n) - P_0(\beta_n)||_{TV}\]
    where $P_0(\beta_n)$ and $P_T(\beta_n)$ are the inferred distributions over agent efficiency $\beta_n$ at the start and time $t$ of the trial $m$ respectively. This captures surprise due to the agent acting more or less efficiently than initially expected. \\
    
    \item \textbf{Change in log observation likelihood}
    \[\max_t \left[\log \frac{P(o_{m,0})}{P(o_{m,t}|o_{m,0:t-1})} \right]\]
    where $P(o_{m,0})$ is the marginal likelihood of the initial observation for trial $m$, and  $P(o_{m,t}|o_{m,0:t-1})$ is the marginal likelihood of the observation at timestep $t$ given all previous observations. This captures surprise due to an observation being inexplicable under the model as a whole, scaled by the likelihood of the initial observation. \\
\end{itemize}

To convert these surprise metrics into a plausibility rating between 0 and 1, we pass each metric $x$ through a logistic regression classifier $f(x)$ with weight $w$ and bias $b$:
\[f(x) = \frac{1}{1 + \exp(wx+b)}\]
The output of each classifier can be viewed as a plausibility rating for each metric, where $p_g$ is the goal plausibility derived from the first metric, $p_\beta$ is the efficiency plausibility derived from the second metric, and $p_o$ is the observational plausibility derived from the third metric. Finally, we multiply these ratings to get the overall plausibility $p$, which can viewed as the probability that the trial is expected. This classifier is depicted in Figure \ref{fig:plausibility-classifier}.

By formulating the plausibility of a trial as a product classifier, we get the intuitive result that if a trial is surprising according to any one metric, then it is surprising overall. For example, if the agent ends up moving towards a goal that is different than initially predicted, the total variation in the goal posterior will be high, leading to a low value of $p_g$, and hence a low plausibility rating $p$. Similarly, if the agent acts very inefficiently after many trials of efficient action, the total variation in the efficiency posterior will be high, leading to a low value of $p_\beta$ and hence $p$.

\section{Results}

\begin{table}[t]
\centering
\begin{tabular}{@{}lllll@{}}
\toprule
\textbf{Task} & \textbf{HBToM} & \textbf{BC-MLP} & \textbf{BC-RNN} & \textbf{Video-RNN} \\ \midrule
Efficiency & \textbf{96.0} & 88.8 & 82.5 & 83.1 \\
\quad \textit{Path Control} & 94.9 & 94.0 & 92.8 & \textbf{99.2} \\
\quad \textit{Time Control} & 97.2 & 99.1 & 99.1 & \textbf{99.9} \\
\quad \textit{Irrational} & \textbf{96.6} & 73.8 & 56.5 & 50.1 \\
Preference & \textbf{99.7} & 26.3 & 48.3 & 47.6 \\
Multi-Agent & \textbf{99.2} & 48.7 & 48.2 & 50.3 \\
Inaccessible Goals & \textbf{99.7} & 76.9 & 81.6 & 74.0 \\
Instrumental Actions & \textbf{98.5} & 67.0 & 77.9 & 79.9 \\
\quad \textit{No Barrier} & 98.8 & 98.8 & 98.8 & \textbf{99.7} \\
\quad \textit{Inconsequential} & \textbf{97.0} & 55.2 & 78.2 & 77.0 \\
\quad \textit{Blocking Barrier} & \textbf{99.7} & 47.1 & 56.8 & 62.9 \\ \bottomrule
\end{tabular}
\caption{Pairwise accuracy of HBToM on BIB vs. baselines.}
\label{tab:results}
\end{table}

We applied the modeling and inference approach described in Section \ref{sec:hbtom} to each BIB episode, resulting in goal and efficiency posteriors for every trial of each episode. From the test trial posteriors, we computed the surprise metrics described in Section \ref{sec:plausibility}, and used them to compute plausibility ratings for the test trial of each episode, where the weights of the logistic classifiers were fine-tuned on synthetic validation dataset (see Appendix). For each set of paired episodes, we compute pairwise correctness by checking if the plausibility rating of the plausible episode is higher than the rating for its implausible counterpart. Averaging across all episode pairs in a task set gives the overall pairwise accuracy for that task.

Table \ref{tab:results} shows the pairwise accuracy of our method (HBToM) on each benchmark task (along with subtasks where present), compared to results from two deep-learning based behavioral cloning baselines (BC-MLP, BC-RNN) and a video prediction RNN baseline (Video-RNN) provided in the original benchmark paper \cite{gandhi2021baby}. As can be seen, our method achieves near-perfect pairwise accuracy on almost all tasks, while significantly outperforming the baseline methods. On a few subtasks, it performs slightly worse than some baselines, but this is made up for by its high performance across the board.

We attribute this success to the structured prior knowledge embedded into our HBToM model, which corresponds closely to an intuitive conceptual understanding of goal-directed agents with preferences over possible goals, and variation in how efficiently they achieve those goals. Because of this prior structure, our method is able to rapidly draw the right inferences about agent dispositions from just a small number of habituation trials, and consequently make human-like judgements about what subsequent behavior is plausible.

\section{Discussion}

Thus far we have focused on the quantitative performance of the HBToM model on BIB. In a future version of this paper, we also aim to present a more detailed qualitative analysis of the many ways in which the HBToM model produces human-like inferences, and uses those inferences to then make plausibility judgements. For example, we hope to illustrate the changes in goal and efficiency posteriors over time, including phenomena such as reversion to the initial goal prior when a new agent is observed, habituation to the inefficiency of an irrational agent, and a lack of change in the posterior over preferred goals when only one goal is accessible.

The multifold aspect of plausibility judgements also lends itself to ablation studies. In the future, we hope to better understand the contribution of each of the surprise metrics, analyzing how well each metric can do on its own without the others. Our hypothesis is that no one metric is sufficient: Total variation in the goal posterior cannot determine plausibility when there is only one goal, and total variation in the efficiency posterior cannot determine plausibility if the agent always acts efficiently (but suddenly changes their goal). Change in log-likelihood might seem like a promising general candidate, but we also expect that inexplicability alone is not sufficient to account for all the ways a human might find an observation surprising. More detailed analysis of our results will allow us to tease apart these factors.

We conclude by reflecting on the limitations and advantages of structured Bayesian models such as our HBToM at the practical and theoretical levels. The primary limitation of such models is that they can require more upfront conceptual and engineering work. Effort is required to design and implement approximately veridical models of environmental dynamics and goal-directed agent behavior, which increases in difficulty with more realistic environments and more complex goals. This is in contrast to deep learning  models trained to mimic human social cognition while making minimal structural assumptions, in which there has been increasing interest \cite{rabinowitz2018machine}.

On the other hand, the performance of deep learning models with minimal inductive bias on both BIB \cite{gandhi2021baby} and the related AGENT benchmark \cite{shu2021agent} suggests that they often fail to generalize in human-like ways, as opposed to Bayesian inverse planning approaches such as our HBToM model and the BIPaCK baseline in \cite{shu2021agent}. The benefits of using such structured models are manifold: They display high accuracy and generalization ability in the environment they are designed for, often with minimal to no training required. Taking advantage of recent advances in probabilistic programming \cite{cusumano2019gen}, they can also be engineered to achieve rapid online inference \cite{zhi2020online}, augmented with neurally-guided inference \cite{cusumano2017probabilistic}, and extended to account for higher-level aspects of human cognition \cite{alanqary2021modeling}.

Perhaps most importantly, structured Bayesian models provide more satisfying \emph{explanations} of human behavior and cognition \cite{lombrozo2006structure}: Model components (e.g. preference priors) and resulting inferences (e.g. goal posteriors) correspond to widely-accepted theoretical constructs and folk psychological concepts (e.g. preferences and goals), while helping us refine fuzzy concepts (e.g. plausibility) in the process of model development. We hope that our HBToM model serves as an illustration of these benefits, inspiring future theoretically-grounded approaches to understanding and replicating human social cognition.

\section*{Acknowledgements}

This work was supported by the DARPA Machine Common Sense program (030523-00001).

%% Use plainnat to work nicely with natbib. 

\bibliographystyle{unsrtnat}
\bibliography{sources}

\pagebreak

\appendix

\section{Appendix}

Here we provide more technical details about our modeling and inference approach, as well as parameter settings for reproducibility. Code is available at \texttt{\url{https://github.com/probcomp/SolvingBIB.jl}}.

\subsection{Agent Priors}

We assume that the preference vector $\theta_n$ has a flat Dirichlet distribution as its prior ($\alpha_i = 1$ for all $i$), a weak prior expressing equal preference for all objects. For the efficiency prior, we assume an inverse Gamma distribution with parameters $a=b=1$, expressing a weak assumption that the agent tends to be efficient rather than random. In practice, the efficiency prior is discretized into a categorical distribution over a fixed set of values ($\beta_n \in \{0.2, 0.2\sqrt{2}, 0.4, 0.4\sqrt{2}, 0.8\}$) for the purposes of rapid and tractable inference.

\subsection{Policy Computation}

Given the relatively small state space of the BIB environment, it is possible to compute the optimal policy $\pi_m(a|s)$ simply via exhaustive value iteration (VI). For greater efficiency however, we instead compute the value function and Q-value function only for those states that occur in the observations, leading to rapid inference that avoids wasteful computation of values and Q-values for unvisited portions of the state-space.

This speed-up is possible because the environment is deterministic, allowing us to directly compute the value function $V(s)$ as the negative of the cost of the cheapest plan from state $s$ to the goal. This cost can be computed by running A* search with an admissible heuristic (e.g. Euclidean distance to the goal) from each visited state $s$ and its neighbors. A* search will produce an optimal plan, and summing the costs of each action in the plan gives the negative of the value function. This can be extended to stochastic domains by using the costs produced by A* search as initial value function estimates in an asynchronous VI algorithm such as Real-Time Dynamic Programming \cite{barto1995learning}, enabling convergence to the true value function and policy without requiring a large number of iterations.

\subsection{Environment States and Observations}

We note that the true environmental dynamics of the BIB environment are continuous. Hence, discretization of the environment results in a loss of precision. However, it also enables efficient policy construction and inference over the discretized state space, avoiding the need to sample continuous motion plans. For similar reasons, we avoid the modelling the gradual disappearance of removable barriers that occurs after they are ``unlocked'' in the instrumental action stimuli. Given that this process is deterministic, we instead assume that removable barriers disappear immediately once the key is placed in the lock, and ignore intermediate observations where the barrier gradually disappears as part of a pre-processing step. We also process the discretized agent trajectories so that zig-zagging motion is smoothed into diagonal motion where possible.

\subsection{Bayesian Inference}

In order to perform rapid and tractable Bayesian inference about agent goals and efficiency parameters, we make use of the following insights and simplifications:

\begin{itemize}
\setlength{\itemsep}{3pt}

\item \textbf{Deterministic observations.}
Since we can reliably determine agent identities $n_{1:m}$, states $\vec s_{m,t}$, and actions $\vec a_{m,t}$ from the observations $\vec o_{m,t}$ (provided as JSON files), we can perform inference as if we are conditioning on those variables directly. Particle filtering extensions can be used to handle noisy observations \cite{zhi2020online}. 

\item \textbf{Agent independence.} The parameters of each agent $n$ are independent. Hence, we can perform inference about each agent separately, ignoring trials with other agents. 
    
\item \textbf{Discretization of efficiency distributions.} Instead of sampling efficiency parameters $\beta_n \sim P(\beta_n)$ to perform approximate inference, we can select a discrete set $\{\beta^1_n, ..., \beta^k_n\}$ to grid over, limiting variance in inference results. By choosing $\beta^1_n, ..., \beta^k_n$ representatively (from efficient to random), we can still draw the desired qualitative inferences about an agent's efficiency.
    
\item \textbf{Goal-preference conjugacy.} Due to conjugacy between the Dirichlet prior $P(\theta_n)$ over agent preferences and the categorical goal distribution $P(g_m|\theta_n)$, we can directly infer the distribution $P(g_m|g_1, .., g_{m-1})$ of the current goal $g_m$ given previous goals $g_{1:m-1}$ using an exact conjugate update. This can be generalized to case where we cannot observe the goals directly. Because this marginalizes out $\theta_n$ exactly, it obviates the need to sample many values of $\theta_n$ to perform inference.
\end{itemize}

Making use of these insights, we perform inference in a trial-by-trial manner, setting the prior over goals for trial $m$ as the goal posterior conditioned on all previous trials $p(g_{m}|\vec o_{m-1,T_{m-1}})$, which can be computed via an exact conjugate update. Inference for the new trial is performed by enumerating over all $k$ values of $\beta_{n_m}$, and computing the probabilities of the actions $a_{m,1:t}$ under the corresponding policy $\pi_{m}$ for each $\beta_{n_m}^i$. Marginalizing over the goal $g_m$ gives the posterior over efficiencies $\beta_{n_m}$, and vice versa.

\subsection{Classifier Tuning}

To tune the weights and biases of the logistic regression classifiers described in Section \ref{sec:plausibility}, we generate a synthetic dataset of 220 episodes similar to the BIB evaluation episodes, but already discretized into the appropriate PDDL representation. (We do not use the BIB training dataset because it lacks implausible examples to tune against). Given the synthetic dataset, we jointly optimize the bias $b$ and weight $w$ associated with the total variation metrics, minimizing cross-entropy loss with L2 regularization of $0.0001$. For change in log-likelihood, we manually tuned the coefficients to account for outliers since synthetic data may not match the test distribution. The resulting weights and biases are shown below:

\begin{itemize}
    \item \textbf{Goal Total Variation}: $w=-11.81$, $b=5.96$
    \item \textbf{Efficiency Total Variation}: $w=-9.42$, $b=5.90$
    \item \textbf{Change in Log. Likelihood}: $w=-0.2$, $b=2.0$
\end{itemize}

\end{document}